\documentclass[preprint,12pt,numbers,sort&compress]{elsarticle}
\usepackage{CJK}   
\usepackage{amssymb}
\usepackage{graphicx}
\usepackage{amsmath}
\usepackage{algorithm}
\usepackage{algpseudocode}
\usepackage{url}
\usepackage{booktabs}
\usepackage{graphicx}
\usepackage{setspace}
\usepackage{varwidth}

\usepackage{amssymb,amsmath}
\usepackage{multirow}
\usepackage{appendix}
\usepackage{changepage}
\usepackage{float}
\usepackage{caption}
\usepackage{bm}
\usepackage{threeparttable}
\usepackage{graphicx}
\usepackage{subfigure}
\usepackage{lineno}
\usepackage{url}
\usepackage{multirow}
\usepackage{booktabs}
\usepackage{amsmath}
\usepackage{amssymb}
\usepackage{bm}
\DeclareMathOperator*{\argmax}{argmax}

\usepackage{dashrule}
\usepackage{setspace}
\usepackage{bibspacing}
\journal{Pattern Recognition}

\begin{document}

\begin{frontmatter}

\title{End-to-End Subtitle Detection and Recognition for Videos in East Asian Languages via CNN Ensemble with Near-Human-Level Performance}
\author[sklsdebhu,msra]{Yan Xu}
\ead{xuyan04@gmail.com}
\author[sklsdebhu]{Siyuan Shan}
\ead{shansiliu@outlook.com}
\author[sklsdebhu]{Ziming Qiu}
\ead{zimingqiubuaa@gmail.com}
\author[msra,qh]{Zhipeng Jia}
\ead{v-zhijia@microsoft.com}
\author[sklsdebhu]{Zhengyang Shen}
\ead{hbgtjxzbbx@gmail.com}
\author[sklsdebhu]{Yipei Wang}
\ead{edithwang525@gmail.com}
\author[bz]{Mengfei Shi}
\ead{shimengfei2012@outlook.com}
\author[msra]{Eric I-Chao Chang\corref{cor1}}
\ead{echang@microsoft.com}
\cortext[cor1]{Corresponding author}

\address[sklsdebhu]{State Key Laboratory of Software Development Environment and Key Laboratory of Biomechanics and Mechanobiology of Ministry of Education and Research Institute of Beihang University in Shenzhen, Beihang University, Beijing 100191, China}
\address[msra]{Microsoft Research Asia, Beijing 100080, China}
\address[qh]{Institute for Interdisciplinary Information Sciences, Tsinghua University, Beijing 100084 , China}
\address[bz]{Beijing No.8 High School, Beijing 100032, China}

\begin{abstract}


\let\thefootnote\relax\footnotetext{\emph{Abbreviations:} STBB, subtitle top/bottom boundary; SLRB, subtitle left/right boundary; CWT, Character Width Transform; SCW, single character width}


In this paper, we propose an innovative end-to-end subtitle detection and recognition system for videos in East Asian languages. Our end-to-end system consists of multiple stages. Subtitles are firstly detected by a novel image operator based on the sequence information of consecutive video frames. Then, an ensemble of Convolutional Neural Networks (CNNs) trained on synthetic data is adopted for detecting and recognizing East Asian characters. Finally, a dynamic programming approach leveraging language models is applied to constitute results of the entire body of text lines. The proposed system achieves average end-to-end accuracies of 98.2\% and 98.3\% on 40 videos in Simplified Chinese and 40 videos in Traditional Chinese respectively, which is a significant outperformance of  other existing methods. The near-perfect accuracy of our system dramatically narrows the gap between human cognitive ability and state-of-the-art algorithms used for such a task.
\end{abstract}

\end{frontmatter}

\section{Introduction}
\label{intro}
Detecting and recognizing video subtitle texts in East Asian languages (e.g. Simplified Chinese, Traditional Chinese, Japanese and Korean) is a challenging task with many promising applications like automatic video retrieval and summarization. Different from traditional printed document OCR, recognizing subtitle texts embedded in videos is complicated by cluttered backgrounds, diversified fonts, loss of resolution and low contrast between texts and backgrounds \cite{Ye2015Text}. 

Given that video subtitles are almost always horizontal, subtitle detection can be partitioned into two steps: subtitle top/bottom boundary (STBB) detection and subtitle left/right boundary (SLRB) detection. These four detected boundaries enclose a bounding box that is likely to contain subtitle texts. Then the texts inside the bounding box are ready to be recognized.

\begin{figure}
\centering
\includegraphics[width=397pt]{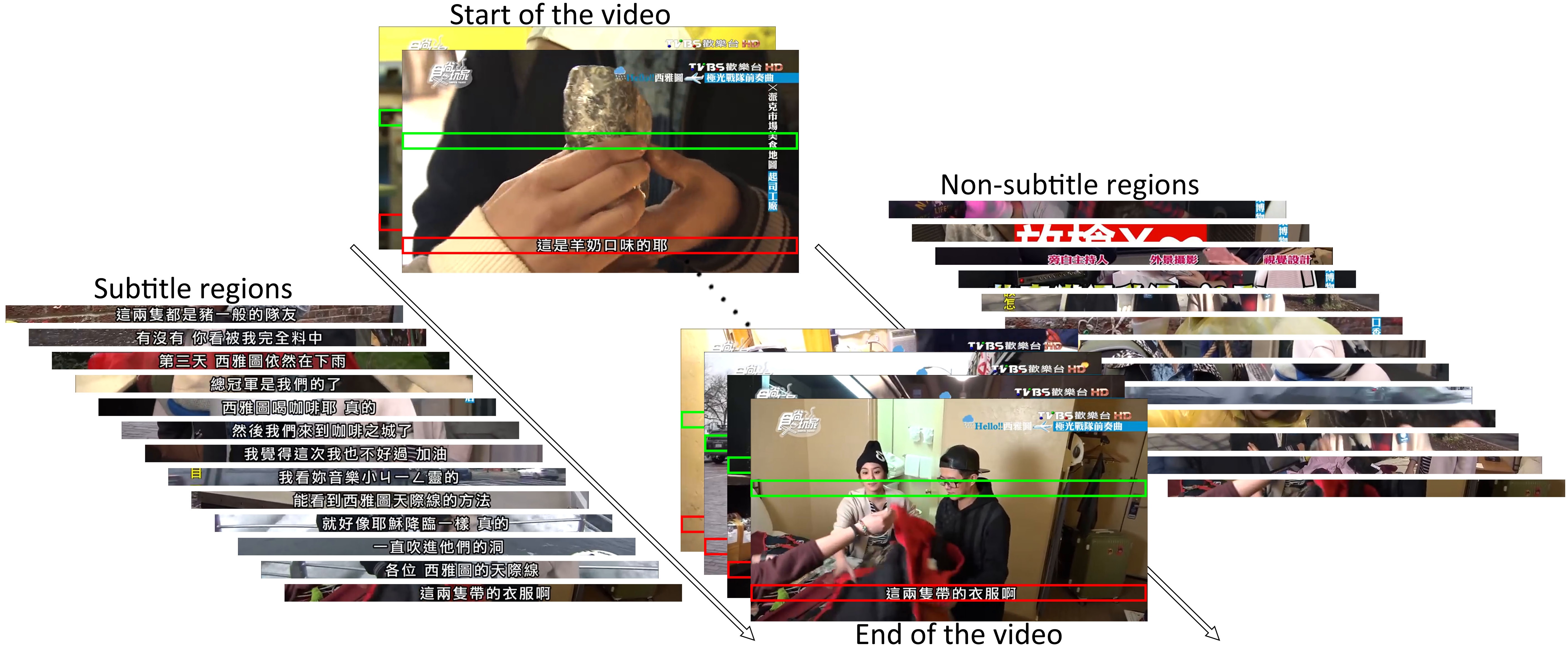}
\caption{Illustration of the consistent STBB position throughout the video. The red box denotes the subtitle region, while the green box denotes the non-subtitle region.} 
\label{motivation}
\end{figure}

Despite the similarity between video subtitle detection and scene text detection (i.e. detect texts embedded in natural static images \cite{Ye2015Text}), the instinctive sequence information of videos makes it necessary to address these two tasks respectively \cite{karatzas2013icdar}. As illustrated in Fig.\ref{motivation}, for most videos with single-line subtitles in East Asian languages, texts at the subtitle region exhibit homogeneous properties throughout the video, including consistent STBB position, color and single character width (SCW). Meanwhile, the non-subtitle region varies unpredictably from frame to frame. With the assistance of this valuable sequence information, we put forward a suitable image operator that can facilitate the detection of STBB and SCW. We call this image operator the \textit{Character Width Transform} (CWT), as it exploits one of the most distinctive features of East Asian characters---consistent SCW.

Considering the complexity of backgrounds and the diversity of subtitle texts, adopting a high-capacity classifier for both text detection and recognition is imperative. CNNs have most recently proven their mettle handling image text detection and recognition \cite{Wang2012End,jaderberg2014deep}. By virtue of their special bio-inspired structures (i.e. local receptive fields, weight sharing and sub-sampling), CNNs are extremely robust to noise, deformation and geometric transformations \cite{rajapakse2012neural} and thus are capable of recognizing characters with diverse fonts and distinguishing texts from cluttered backgrounds. Besides, the architecture of CNNs enables efficient feature sharing across different tasks: features extracted from hidden layers of a CNN character classifier can also be used for text detection \cite{jaderberg2014deep}. Additionally, the fixed input size of typical CNNs makes them especially suitable for recognizing East Asian characters whose SCW is consistent.

In view of the straightforward generation pipeline of video subtitles, it is technically feasible to obtain training data by simulating and recovering this generation pipeline. To be more specific, when equipped with a comprehensive dictionary, several fonts and numerous random backgrounds, machines can produce huge volumes of synthetic data covering thousands of characters in diverse fonts without strenuous manual labeling. As a cornucopia of synthetic training data meet the ``data-hungry'' nature of CNNs, models trained merely on synthetic data can achieve competitive performance on real-world datasets.

Another observation is that the recognition performance degrades with the burgeoning number of character categories (as in the case of East Asian languages). In a similar circumstance, Jaderberg et al. \cite{jaderberg2016reading} attempt to alleviate this problem with a sophisticated incremental learning method. Here we propose a more straightforward solution: instead of using a single CNN, we independently train multiple (ten in this paper) CNN models that consolidate a CNN ensemble. These models are complementary to each other, as the training data is shuffled respectively for training different models.

In this paper, by seamlessly integrating the above-mentioned cornerstones, we propose an end-to-end subtitle text detection and recognition system specifically customized to videos with a large concentration of subtitles in East Asian languages. 
Firstly, STBB and SCW are detected based on a novel image operator with the sequence information of videos. SCW being determined at an early stage can provide instructive information to improve the performance of the remaining modules in the system.  Afterwards, SLRB is detected by a SVM text/non-text classifier (it takes CNN features as input) and a horizontal sliding window (its width is set to SCW). According to the detected top, bottom, left and right boundaries, the video subtitle is successfully detected. Finally, single characters are recognized by the CNN ensemble and the text line recognition result is determined by a dynamic programming algorithm leveraging a 3-gram language model. We show that the CNN ensemble produces a recognition accuracy of 99.4\% on a large real-world dataset including around 177,000 characters in 20,000 frames. This dataset with ground truth annotations and our CNN models will be made publicly available.

Our contribution can be summarized as follows:

\begin{itemize}

\item We propose an end-to-end subtitle detection and recognition system for East Asian languages. By achieving 98.2\% and 98.3\% end-to-end recognition accuracies for Simplified Chinese and Traditional Chinese respectively, this system remarkably narrows the gap to human-level reading performance\footnote{Human-level reading performance is 99.6\% according to the experiment in Section \ref{human}.}.

\item We define a novel image operator whose outputs enable the effective detection of STBB and SCW. The sequence information is integrated throughout the video to increase the reliability of the proposed image operator. This module achieves a competitive result on a dataset including 1,097 videos.

\item We leverage a CNN ensemble to perform the classification of East Asian characters across huge dictionaries. The ensemble reduces the recognition error rate by approximately 75\% in comparison with a single CNN. CNNs in our system serve both as text detectors and character recognizers owing to efficient feature sharing. The visualization of CNNs proves that different CNN models can capture distinctive features of characters.

\end{itemize}

The remainder of this paper is organized as follows. Section \ref{relatedwork} reviews related works. Section \ref{method} describes the synthetic data generation scheme, the CNN ensemble and the end-to-end system. In Section \ref{expr}, the proposed system and each module in it are evaluated on a large dataset, and the experimental results are presented. In Section \ref{discuss}, observations from our experiments are discussed. A conclusion and discussion of future work are given in Section \ref{conc}.

\section{Related work}
\label{relatedwork}
In this section, we focus on reviewing relevant literature on image text detection and recognition. As for other text detection and recognition methods, several review papers \cite{Ye2015Text,Jung2004Text,Sharma2012Recent,Zhang2008Extraction} can be referred to. 

\subsection{Image text detection}
\label{textdetection}
Generally, text detection methods are based on either connected components or sliding windows \cite{jaderberg2014deep}. Connected component based methods, like Maximally Stable Extremal Regions (MSER) \cite{Matas2004Robust,Shi2013Scene,huang2014robust}, enjoy their computational efficiency and high recall rates, but suffer from a large number of false detections. Methods based on sliding windows \cite{coates2011text,Wang2012End,jaderberg2014deep,Kai2011End,Delakis2008text,Ren2015A} adopt a multi-scale window to scan through all locations of an image, then apply a trained classifier with either hand-engineered features or learned features to distinguish texts from non-texts. Though this kind of method produces significantly less false detections, the computational cost of scanning every location of the image is unbearable. Therefore, connected component based methods and sliding-window based methods are often utilized together for text detection \cite{Alsharif2013End,jaderberg2016reading,neumann2010method,huang2014robust}, where the former generate text region proposals and the latter eliminate false detections. This text detection scheme is also adopted in this paper, but our text region proposal method is based on the sequence information of video and thus not comparable to existing methods designed for scene text detection. Hence, we focus on reviewing methods based on video sequence information and text region verification works that aim to eliminate false detections.

\subsubsection{Methods incorporating video sequence information}

Tang et al. \cite{tang2002spatial} analyze the difference of adjacent frames to detect the subtitle text based on the assumption that in each shot the scene changes more gradually than the subtitle text. Wang et al. \cite{wang2004novel} exploit a multi-frame integration technique within 30 consecutive frames to reduce the complexity of backgrounds before the text detection process. Liu et al. \cite{liu2012robustly} compare the distribution of stroke-like edges between adjacent frames and segment the video into clips in which the same caption is contained. Then they adopt a temporal ``and'' operation to identify caption regions. However, contrary to the proposed method in this paper, these existing methods rarely exploit temporal information throughout the video.

\subsubsection{Text region verification based on hand-engineered features}
Traditional methods harness manually designed low-level features such as SIFT and histogram of oriented gradients (HOG) to train a classifier to distinguish texts from non-texts. For instance, Wang et al. \cite{wang2009new} propose a new block partition method and combine the edge orient histogram feature with the gray scale contrast feature (EOH-GSC) for text verification. Neumann et al. \cite{neumann2010method} adopt the SVM classifier with a set of geometric features for text detection. Wang et al. \cite{Kai2011End} and Jaderberg et al. \cite{jaderberg2016reading} eliminate false text detections by Random Ferns with HOG features. Minetto et al. \cite{Minetto2013T} propose a HOG-based texture descriptor (T-HOG) that ameliorates traditional HOG features on the text/non-text discrimination task. Effective as these handcrafted features are to describe image content information, they are suboptimal to represent text data due to their heavy dependence on priori knowledge and heuristic rules.

\subsubsection{Text region verification based on feature learning}
In contrast to these traditional methods, more advanced methods take advantage of high-capability  feature learning to automatically learn a more robust representation of text data,  hence possessing a  powerful discrimination ability to eliminate false text detections. Delakis and Garcia \cite{Delakis2008text} train a CNN to detect texts from raw images in a sliding window fashion. Wang et al.  \cite{Wang2012End} and Huang et al. \cite{huang2014robust} utilize a multi-layer CNN for both text detection and recognition, and the first layer of the network is trained with an unsupervised learning algorithm \cite{coates2011text}. Ren et al. \cite{Ren2015A} are the first to tackle Simplified Chinese scene text detection. They propose an algorithm called convolutional sparse auto-encoder (CSAE) to pre-train the first layer of CNN on unlabeled synthetic data for Simplified Chinese scene text detection.

Both the above-mentioned methods and our approach are based on feature learning, comparing favorably against methods based on hand-engineered features. We further promote East Asian text detection performance by training a CNN ensemble in an end-to-end manner on labeled synthetic data.

\subsection{Image text recognition}
Similar to Section \ref{textdetection} where the importance of features is addressed, existing image text recognition methods are also classified into those based on hand-engineered features \cite{lee2014region,wang2010word,Bissacco2013PhotoOCR,neumann2010method,Kai2011End,Bai2014Chinese2} and those based on feature learning \cite{coates2011text,Alsharif2013End,Wang2012End,jaderberg2014deep,jaderberg2016reading,Saidane2007Automatic,Saidane2009The,netzer2011reading,bai2014image,elagouni2014text,Zhong2015Multi,Elagouni2011A}.

\subsubsection{Image text recognition based on hand-engineered features}
Bissacco et al. \cite{Bissacco2013PhotoOCR} propose a scene text recognition system by combining a neural network trained on HOG features with a powerful language model. Lee et al. \cite{lee2014region} present a new text recognition method by merging gradient histograms, gradient magnitude and color features. Bai et al. \cite{Bai2014Chinese2} use HOG features, artificially generated training data and a neural network classifier for Simplified Chinese image text recognition. Though state-of-the-art performance was achieved, its 85.44\% recognition accuracy still impedes its practical application.
\subsubsection{Image text recognition based on feature learning}
Elagouni et al. \cite{Elagouni2011A} harness a CNN to perform character recognition with the aid of a language model, and their system achieves outstanding performance on 12 videos in French. Jaderberg et al. \cite{jaderberg2014deep} propose a novel CNN architecture that facilitates efficient feature sharing for different tasks like text detection, character classification and bigram classification. Alsharif and Pineau \cite{Alsharif2013End} utilize the Maxout network \cite{Goodfellow2013Maxout} together with an HMM with a fixed lexicon to recognize image words. Jaderberg et al. \cite{jaderberg2016reading} propose a CNN that directly takes whole word images as input and classifies them across a dictionary of 90,000 English words.

Works tackling East Asian image text recognition with CNNs are relatively rare. Zhong et al. \cite{Zhong2015Multi} adopt a CNN with a multi-pooling layer on top of the final convolutional layer to perform multi-font printed Simplified Chinese character recognition, which renders their method robust to spatial layout variations and deformations. Bai et al. \cite{bai2014image} propose a CNN architecture for Simplified Chinese and English character recognition, and the hidden-layers are shared across these two languages. However, both  works \cite{Zhong2015Multi,bai2014image} can only recognize an isolated character as opposed to a text line. Besides, the work of Bai et al. \cite{bai2014image} can only recognize 500 Simplified Chinese characters, though there are thousands of characters commonly used \cite{wang2001optical}. Therefore, to the best of our knowledge, the system proposed in this paper is the first to leverage high-capability CNNs to recognize image text lines in Simplified Chinese (and also other East Asian languages) with a comprehensive alphabet consisting of 7,008 characters.

\section{Method}
\label{method}
\begin{figure}[t]
\centering
\includegraphics[width=255pt]{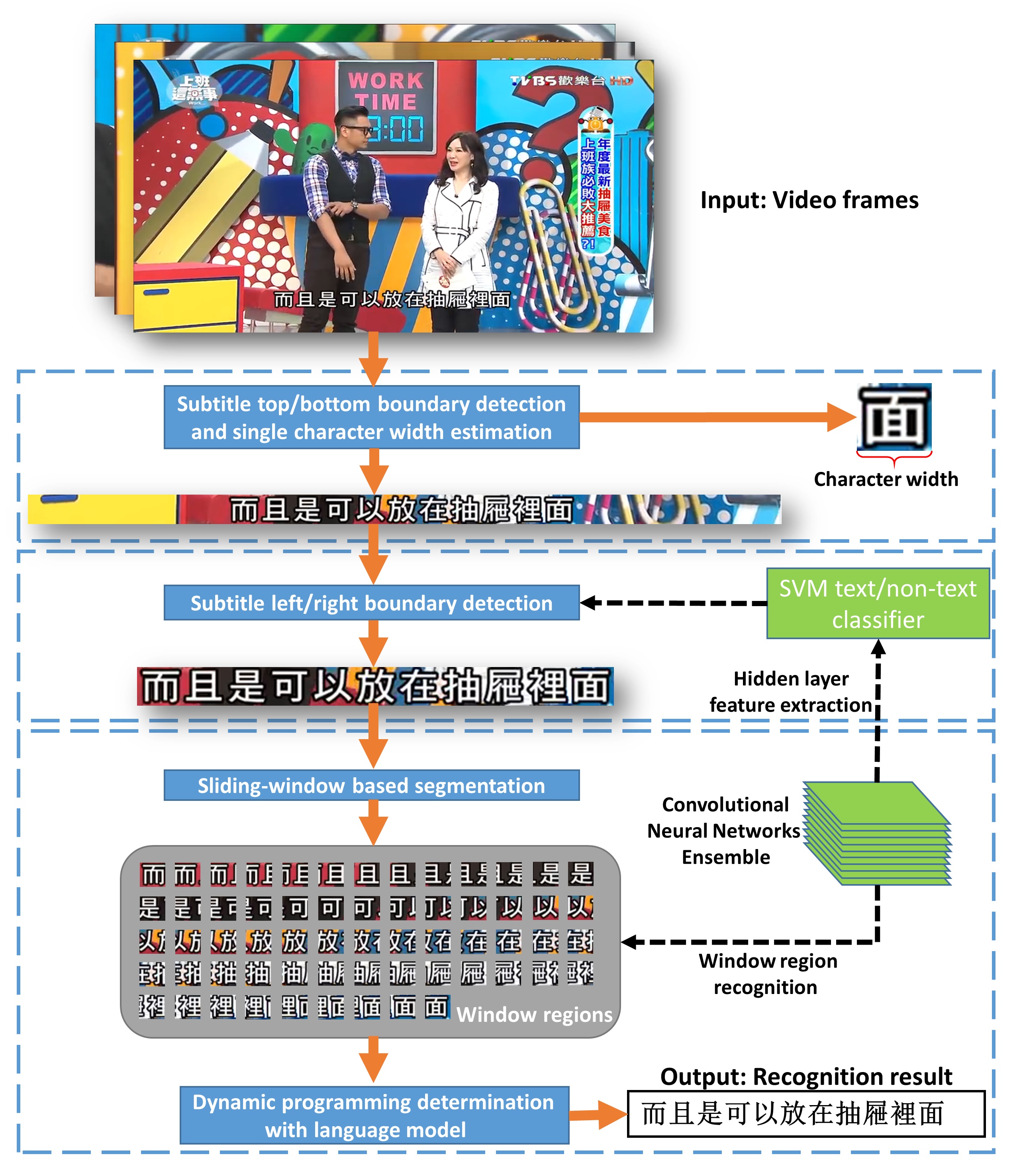}
\caption{Overview of the proposed system. The end-to-end system consists of three modules corresponding to three boxes with blue dashed borders in the figure. Given  a set of video frames, the first module detects STBB and SCW. In the second module, SLRB is detected by a SVM text/non-text classifier with features extracted from the hidden layer of the CNN ensemble. In the third module, a sliding window with width equaling to SCW is employed, and the CNN ensemble recognizes characters in each window region. The final result is given by a dynamic programming algorithm with a language model.}
\label{pipeline}
\end{figure}

In this section, we will describe the synthetic data generation pipeline, the CNN ensemble and the end-to-end system in detail. As illustrated in Fig.\ref{pipeline}, the end-to-end system consists of three modules including STBB and SCW detection, SLRB detection and subtitle recognition.

\subsection{Synthetic data generation}
\label{synthetic_data}
As it is easy to simulate the generation pipeline of subtitles, training data are synthetically generated in a scheme similar to \cite{bai2014chinese,Jaderberg2014Synthetic}. The labeled synthetic data in Simplified Chinese (SC), Traditional Chinese (TC) and Japanese (JP) are generated to train CNNs in SC, TC and JP respectively. 

(1) Dictionary construction: three comprehensive dictionaries that respectively cover 7,009 SC characters, 4,809 TC characters and 2,282 JP characters are constructed. A space character is  included in each dictionary. 

(2) Font rendering: 22, 19 and 17 kinds of font for SC, TC and JP are collected respectively for introducing more variations to the training data.

(3) Random selection of background and character: 45,441 frames are randomly extracted from 11 news videos downloaded from the Internet. Afterwards, small background patches are randomly cropped from these frames. The size of every background patch is determined with regard to a random combination of a character and a font. 200,000 machine-born white characters with dark shadows are generated by repeatedly selecting a random combination of a font and a character from the dictionary.

(4) Random shift and Gaussian blur: every randomly generated machine-born character is superimposed on a randomly selected background patch with a random shift of $\theta$ pixels, where $\theta$ is drawn from a uniform distribution on the interval [-2, 2]. Then every image is convolved with a Gaussian blur at the scale of $\sigma$ pixels, where $\sigma$ is drawn from a uniform distribution on the interval [0.5, 1.6]. The convolved images are then converted to grayscale images and resized to 24 $\times$ 24. Therefore, 200,000 samples are generated for SC, TC, and JP respectively.

\begin{figure}
\centering
\includegraphics[width=255pt]{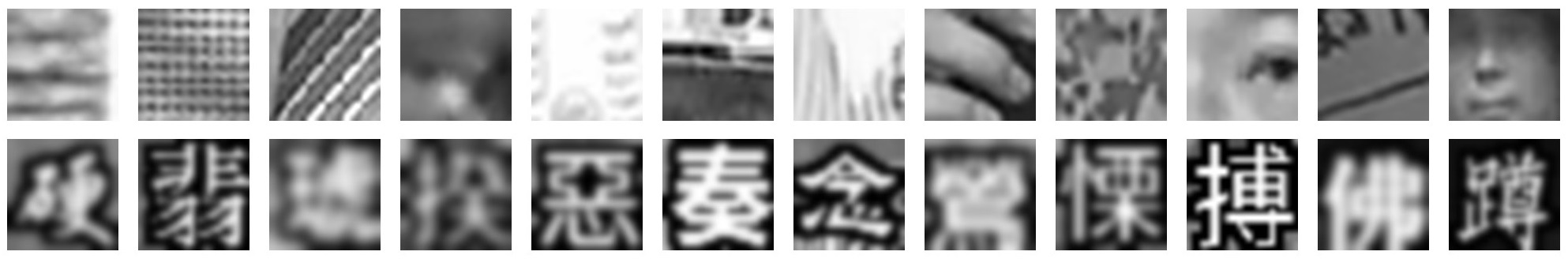}
\caption{Examples of the machine-simulated training data. The small patches on the first line are non-text training examples, while those on the second line are text training examples.}
\label{trainingdata}
\end{figure}

The procedure of generating training samples for the text/non-text SVM classifier is almost the same, except that the same number of background patches without characters are also stored as non-text training examples.
Fig.\ref{trainingdata} presents some of the training data.

\subsection{Convolutional Neural Networks ensemble}
\label{section_CNN}
CNNs have been recently applied to recognize image texts with great success \cite{jaderberg2014deep,jaderberg2016reading,Alsharif2013End,Wang2012End}. The architecture of our CNN model is mainly inspired by \cite{krizhevsky2012imagenet}, in which a four-layer CNN with local response normalization achieved an 11\% test error rate on the \emph{CIFAR-10} dataset \cite{krizhevsky2009learning}. As delineated by Table \ref{CNN_config}, the configuration of our net is derived from the code shared by Krizhevsky \cite{code_shared}. Our CNN takes as input a character image rescaled to the size of 24 $\times$ 24 pixels and returns as output a vector of $z$ values between 0 and 1. The input image is converted to grayscale image so as to reduce the susceptibility of our model to variable text colors and alleviate the computational burden.

\begin{table}[tbp]
\centering
\footnotesize
\begin{tabular}{@{}lllll@{}}
\toprule
Layer  & Type                & Size-in                & Size-out               & Kernel                 \\ \midrule
conv1  & convolutional       & 24$\times$24$\times$1  & 24$\times$24$\times$64 & 5$\times$5$\times$64,1  \\
pool1  & max-pooling         & 24$\times$24$\times$64 & 12$\times$12$\times$64 & 3$\times$3$\times$64,2 \\
rnorm1 & local response norm & 12$\times$12$\times$64 & 12$\times$12$\times$64 &                        \\
conv2  & convolutional       & 12$\times$12$\times$64 & 12$\times$12$\times$64 & 5$\times$5$\times$64,1 \\
rnorm2 & local response norm & 12$\times$12$\times$64 & 12$\times$12$\times$64 &                        \\
pool2  & max-pooling         & 12$\times$12$\times$64 & 6$\times$6$\times$64   & 3$\times$3$\times$64,2 \\
local3 & locally-connected   & 6$\times$6$\times$64   & 6$\times$6$\times$64   & 3$\times$3$\times$64,1 \\
local4 & locally-connected   & 6$\times$6$\times$64   & 6$\times$6$\times$32   & 3$\times$3$\times$32,1 \\
fc     & fully-connected     & 6$\times$6$\times$32   & $z$                    &                        \\
probs  & softmax             & $z$                    & $z$                    &                        \\ \bottomrule
\end{tabular}
\caption{ CNN configuration. The input and output sizes are described in $rows\times cols\times \# channels$. The kernel is specified as $rows\times cols\times \# filters, stride$. $z$ represents number of character categories.}
\label{CNN_config}
\end{table}

Note that we do not perform the data augmentation as proposed by \cite{krizhevsky2012imagenet}, in which 24 $\times$ 24 patches are randomly cropped from the original 32 $\times$ 32 images in \emph{CIFAR-10} \cite{krizhevsky2009learning} to prohibit overfitting. The reason behind this is twofold. On the one hand, the loss of critical information, including radicals and strokes in characters, is inevitable if the original images are randomly cropped. On the other hand, we are not concerned about overfitting because our synthetic dataset can be arbitrarily large.


\subsubsection{Details of learning}

Stochastic gradient descent with a batch size of 128 images is used to train our models. Parameters like learning rates, weight decay and momentum are concurrent with the shared code \cite{layer_para}. 195,000 images are used for training while the remaining 5,000 images are used for validation. We train each model for only one epoch on the training set, which takes approximately two hours on one NVIDIA Tesla K20Xm GPU.

\subsubsection{Visualization}
\begin{figure}[t]
\centering
\includegraphics[width=255pt]{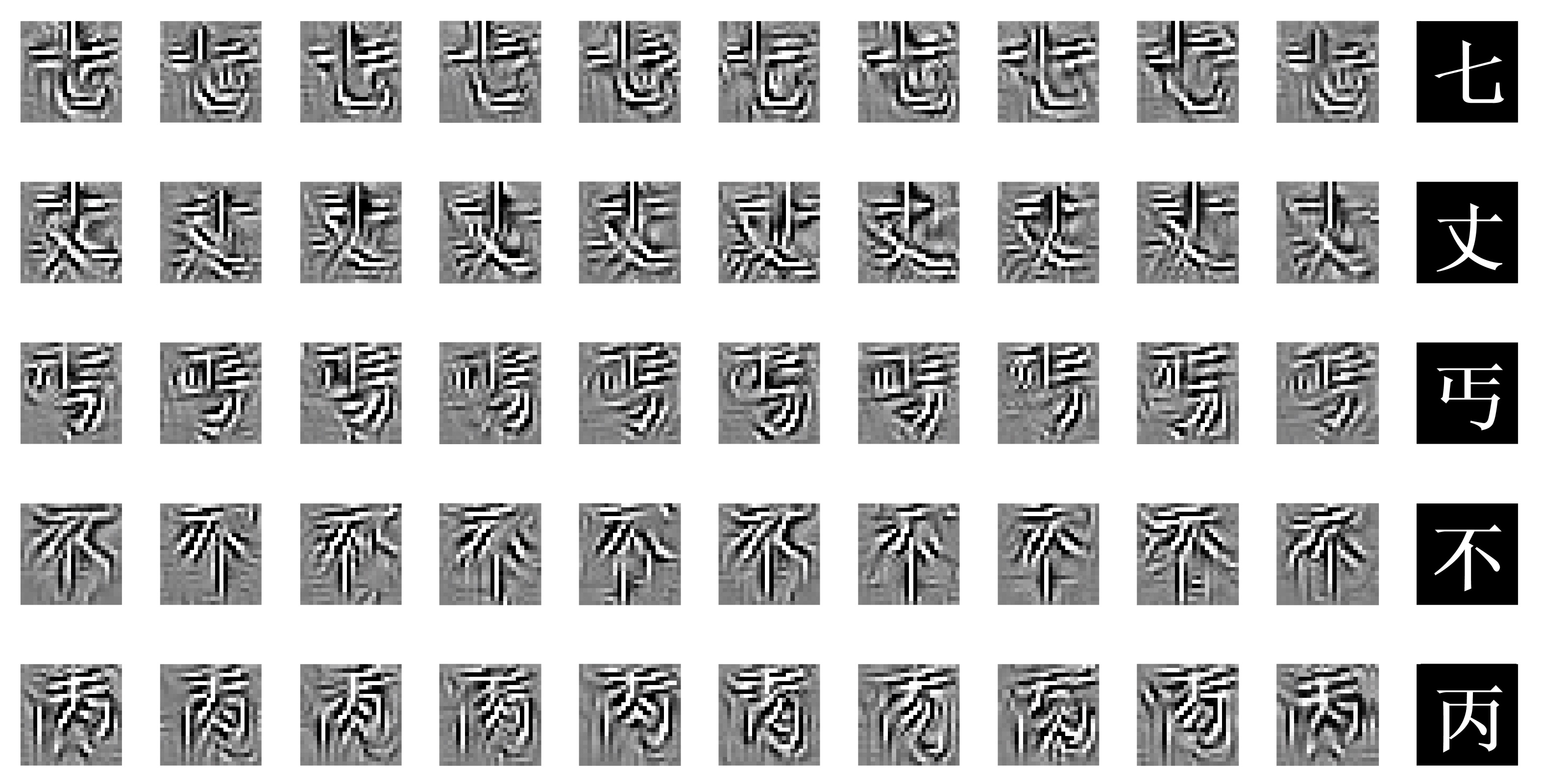}
\caption{Visualization of 5 character classes learned from the Traditional Chinese character classifier. There are 10 visualization results corresponding to 10 CNN models in each line. These images are generated by numerically optimizing the input image which maximizes the score of a specific character category \cite{simonyan2013deep,erhan2009visualizing}.}
\label{visualization}
\end{figure}

In Fig.\ref{visualization}, we visualize the learned CNN ensemble using the technique as demonstrated \cite{simonyan2013deep,erhan2009visualizing}. It can be observed that the appearance of different shifts and fonts of a specific category is captured in a single image, and ten CNN models in the CNN ensemble learn something slightly different from each other albeit the overall similarity. The visualization indicates that the CNN ensemble has captured distinctive features of characters.

\subsubsection{Training the text/non-text SVM classifier}
\label{SVM}
We adopt a linear SVM classifier \cite{cortes1995support} to determine whether there is a character in a given image patch. The SVM takes the outputs of the \emph{local4} layer of the CNN ensemble as its features. The \emph{local4} layer of every CNN outputs a 6 $\times$ 6 $\times$ 32 feature map, which is 1152-dimensional after concatenation. The CNN ensemble consists of 10 CNNs, thus the feature vector of the SVM is 11520-dimensional. The parameter $C$ of the SVM controls the trade off between margin maximization and errors of the SVM on training data. $C$ is optimized on the synthetic validation set.


\subsection{STBB and SCW detection}


In this section, we describe the proposed image operator CWT and how it is applied with the sequence information to detect STBB and SCW.


\subsubsection{Character Width Transform}
\label{CWT}
One feature that distinguishes East Asian text from other elements of a video frame is its consistent SCW. SCWs of East Asian characters are identical as long as their font styles and font sizes are set the same. In this work, we leverage this fact to define CWT, which recovers regions that are likely to contain texts.

\begin{figure}
\centering
\includegraphics[width=397pt]{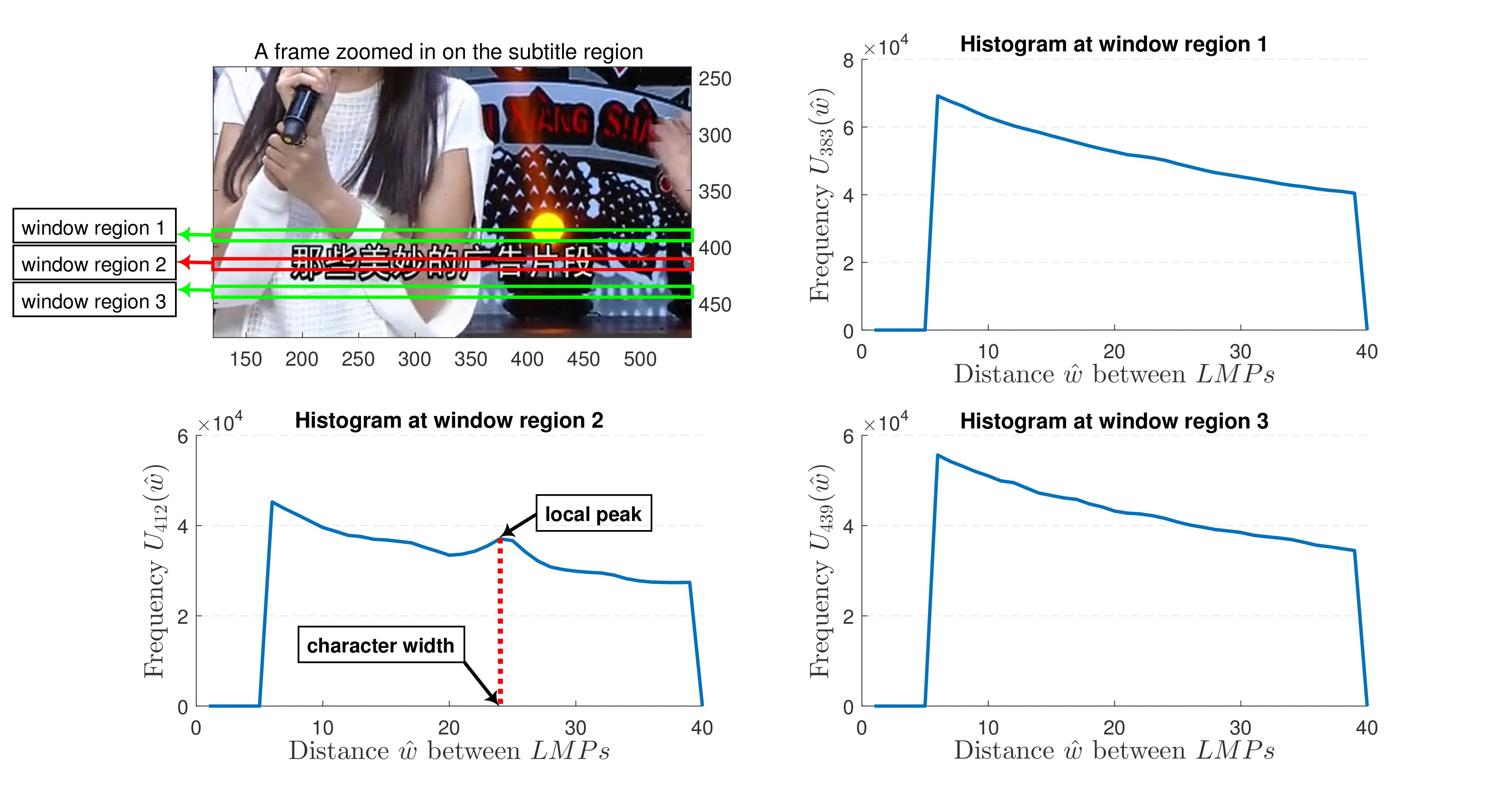}
\caption{Illustration of the distribution patterns of histograms at a subtitle region (window region 2) and non-subtitle regions (window region 1 and 3).}
\label{hist_norm}
\end{figure}

CWT is a local image operator. At each local region, CWT generates a histogram that estimates the distribution of SCWs of the subtitle text in this region. SCW is estimated by detecting pixels that are likely to locate at the space between characters and calculating the pairwise distances between these detected pixels. As illustrated in Fig.\ref{hist_norm}, the randomness at non-subtitle regions makes the pairwise distances distribute uniformly. Meanwhile, at subtitle regions, more pairwise distances come from the space between characters, leading to the emergence of a local peak in the vicinity of the SCW. Based on the distribution patterns of histograms constructed at different local regions, we predicate that the STBB and the SCW can be determined simultaneously.

\begin{figure}
\centering
\includegraphics[width=397pt]{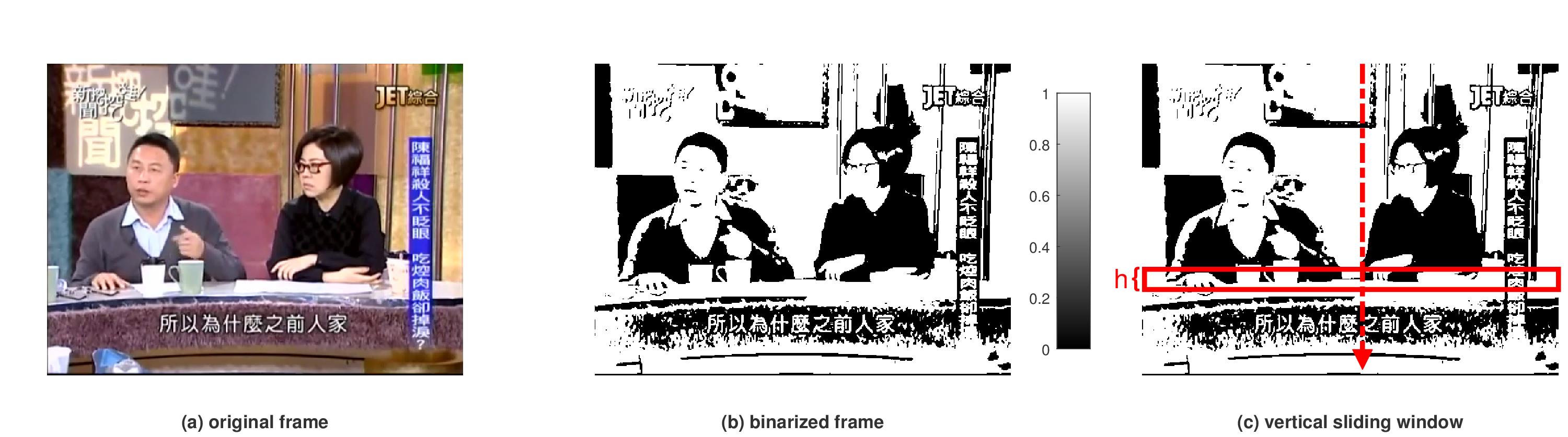}
\caption{(a) is an original RGB frame and (b) is the binarized frame. (c) illustrates the proposed vertical sliding window. In (c), the red box represents the vertical sliding window, and the dashed red arrow shows the direction in which the sliding window moves.}
\label{prepro}
\end{figure}

Detecting pixels at the space between characters requires the binarization of frames extracted from videos (see Fig.\ref{prepro} (b) for illustration). Firstly, each RGB frame with the size of $H\ \times \ W$ is transformed into LAB color space to avoid the illumination inference \cite{Qu2013Hierarchical}. Then, Sauvola algorithm \cite{Sauvola2000Adaptive} is adopted to separate text components from background (binarization) for its robustness to the uneven illumination and noise. This algorithm performs local thresholding with $\mu$-by-$\nu$ neighborhood. Both $\mu$ and $\nu$ are set to 150 pixels and the threshold is set to 0.34.

CWT is then applied to every local region in a sliding-window manner. Concretely, a ${h\ \times \ W}$ sliding window (as shown in Fig.\ref{prepro} (c)) is adopted, where $h$ is a variable less than $H$ and determined according to the resolution of videos. This window scans each frame by moving vertically from top to bottom at stride 1, and $H-h+1$ window regions can be obtained. Finally we acquire $H-h+1$ histograms by applying CWT at every window region.

\begin{figure}[htb]
\centering
\includegraphics[width=397pt]{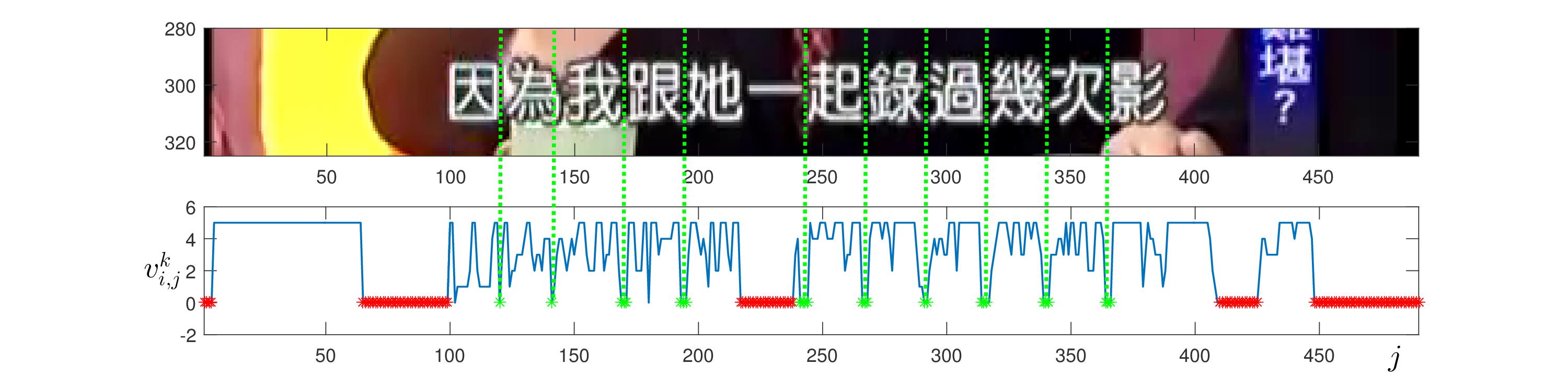}
\caption{The majority of $LPMs$ are interspersed among backgrounds (denoted by red asterisks) and the space between characters (denoted by green asterisks).}
\label{lpms}
\end{figure}

Let $x_{i,j}^{k}\in \left \{ 0,1 \right \}$ denote a pixel in the binarized frame $k$ where $(i,j)$ are the coordinates. Values of most text pixels are 1 after the binarization. We take the sliding-window region whose top boundary is at position $i$, and the sum of elements in its each column is:
\begin{equation}
\label{sum_column}
v_{i,j}^{k}=\sum_{r=i}^{i+h-1}x_{r,j}^{k},
\end{equation}
After that, pixels that are likely to locate at the space between characters are detected by local-minimum points ($LMPs$). We denote a set of $LMPs$ by $\mathcal{L}_{i}^{k}$, where $\mathcal{L}_{i}^{k}=\left \{ x_{i,j}^{k}\ \biggm|\ v_{i,j}^{k}<{\rm min}(v_{i,j-1}^{k},v_{i,j+1}^{k})\   {\rm or}\ v_{i,j}^{k}=0 \right \}$. As illustrated by Fig.\ref{lpms}, the majority of $LMPs$ are interspersed among backgrounds as well as the space between characters. If more than 30 $LMPs$ are connected (i.e. $\forall j$, $\exists M\geq30$, $x_{i,j}^{k}$, $x_{i,j+1}^{k}\dots x_{i,j+M-1}^{k}\in \mathcal{L}_{i}^{k}$), they will be removed, which can effectively eliminate $LMPs$ from backgrounds while reserve $LMPs$ from the space between characters. The rationality of this constraint is that more than 30 connected $LMPs$ could only come from backgrounds.
Then all pairwise distances between $LMPs$ are calculated and stored in a set $\mathcal{D}_{i}^{k}$:
\begin{equation}
\label{dis}
\mathcal{D}_{i}^{k}=\left \{ \left | m-n \right |\ \biggm|\ x_{i,m}^{k},x_{i,n}^{k}\in \mathcal{L}_{i}^{k},\ w_{min}<\left | m-n \right |<w_{max} \right \},
\end{equation}
where $w_{min}$ and $w_{max}$ denote the minimum and the maximum SCW respectively.

It is noteworthy that since the statistical information derived from a single frame is too coarse to provide a reliable estimation of SCW, we can not construct a histogram directly from $\mathcal{D}_{i}^{k}$ in the next step. This is when the sequence information of video comes in handy. As STBB and SCW are consistent throughout the video, we assume that values in $\mathcal{D}_{i}^{1}, \mathcal{D}_{i}^{2}\dots \mathcal{D}_{i}^{T}$ are drawn from the same underlying distribution, where $T$ represents the number of frames in the video. Based on this assumption, histograms $U_{i}\left (\hat{w} \right)$ can be constructed from frames throughout the video:

\begin{equation}
\label{constructU}
U_{i}\left (\hat{w} \right)=\sum_{k=1}^{T}\sum_{r\in \mathcal{D}_{i}^{k}}\textbf{1}_{\hat{w}}\left ( r \right ),
\end{equation}
where $\textbf{1}_{\hat{w}}\left ( r \right )$ equals 1 if $r=\hat{w}$ and 0 otherwise. In order to alleviate the computational burden, videos are downsampled to 0.0625 fps without compromising the STBB detection performance.

\subsubsection{Detecting the STBB and SCW}
Given histograms $U_{1}$,$U_{2}$$\dots$$U_{H-h+1}$, the STBB and the SCW can be determined. Concretely, if the local peaks (see Fig.\ref{hist_norm}) of several adjacent histograms $U_{t}$,$U_{t+1}$$\dots$$U_{b}$ all locate near $\hat{w_{0}}$, $t$ and $b$ will be regarded as positions of a set of candidate STBB, and $\hat{w_{0}}$ will be the corresponding SCW. Our algorithm is presented in Algorithm \ref{algorithm1}, of which the output $\mathcal{P}$ contains several candidate sets of STBB and estimated SCW.

 \begin{algorithm}
\footnotesize
	\caption{STBB and SCW determination}
	\label{algorithm1}
	\begin{algorithmic}[1]
	    \Require \begin{varwidth}[t]{\linewidth}
	        histograms $\{U_{1}, U_{2}\dots,U_{H-h+1}\}$,\par
	        maximum SCW $w_{max}$, minimum SCW $w_{min}$,\par
	        minimum subtitle height $min\_height$
	    \end{varwidth}
	    \Statex
	    \Ensure candidate STBB and SCW $\{\mathcal{P}\}$
	    
	    \Statex
	    
%

        \Statex \emph{Find local peaks inside histograms:}
		\For {$i\gets 1$ \textbf{to} $H-h+1$}
		    \For {$j\gets w_{min}$ \textbf{to} $w_{max}$}
		        \State $q_{i,j}\gets 0$
		        \If {$\max(U_{i}\left ( j-1 \right ),U_{i}\left ( j+1 \right )) \leq U_{i}\left ( j \right ) $} 
		            \State \begin{varwidth}[t]{\linewidth}
		                Estimate the position of local peak by quadratic interpolation as \par
		                \hskip\algorithmicindent $q_{i,j}\gets j+\frac{1}{2}\times \frac{U_{i}\left ( j-1 \right )-U_{i}\left ( j+1 \right )}{U_{i}\left ( j-1 \right )-2\times U_{i}\left ( j \right )+U_{i}\left ( j+1 \right )}$
		            \end{varwidth}
                \EndIf
            \EndFor
        \EndFor
        \Statex
        \Statex \emph{Detect adjacent histograms with similar local peak positions:}
       \State $Q\gets\varnothing,\ \mathcal{P}\gets\varnothing$
       \For {$i\gets 1$ \textbf{to} $H-h+1$}
            \For {$j\gets w_{min}$ \textbf{to} $w_{max}$}
                \If {$q_{i,j}>0$}
                    \State $Q\gets Q\bigcup q_{i,j}$
                    \For {$k\gets i+1$ \textbf{to} $H-h+1$}
                        \State $C\gets \left \{ x\ |\ x\in \left \{ q_{k,j-1},q_{k,j},q_{k,j+1}  \right \},\ x> 0 \right \}$
                        \If {$C=\varnothing$}
                            \State \textbf{break for}
                        \EndIf
                        \State $e\gets \argmax_{x\in C}|x-{\rm median}(Q)|$
                        \State $Q\gets Q \bigcup e$
                    \EndFor
                    \If {$k-i+\lfloor h/2 \rfloor +1\geq min\_height$}
                        \State $\mathcal{P}\gets \mathcal{P}\ \bigcup \ (i,\ k+\lfloor h/2\rfloor +1,\ \left \lfloor {\rm median}(Q)\right \rfloor)$
                    \EndIf
                \EndIf
            \EndFor
       \EndFor

	\end{algorithmic}
 \end{algorithm}

Note that elements contained in $\mathcal{P}$ are raw candidates, some of which might come from non-subtitle regions and should be eliminated. A post processing algorithm are adopted to remove these false-positive candidates: (1) if two candidates with a similar SCW are overlapped, we eliminate the one whose subtitle height is smaller. (2)	if two candidates have a similar STBB and the SCW of one of them is approximately two times larger than the other one, the candidate with the larger SCW is eliminated. (3) candidates whose STBB locate at the upper half of the frame are eliminated due to the fact that most of subtitles are superimposed on the bottom half of the frame.

This post processing algorithm eliminates almost all false detections, and a small amount of surviving false-positives will be further removed by the text/non-text classifier in the step following.

\subsection{ SLRB detection}
\label{SLRB}
\begin{figure}
\centering
\includegraphics[width=397pt]{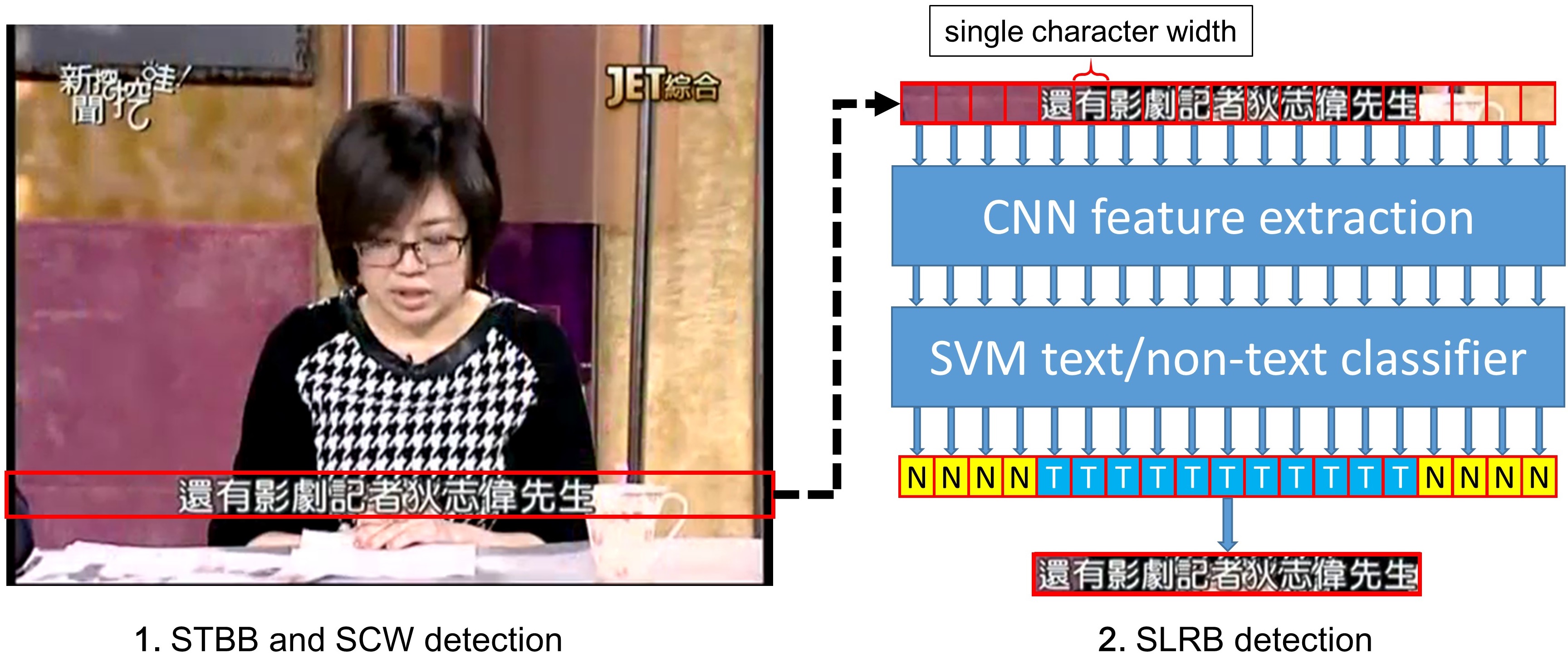}
\caption{This delineates the subtitle detection procedure. STBB and SCW are detected firstly. Then a sliding window horizontally scans the subtitle region detected in the first step. Every window region is predicted either as text (T) or non-text (N) by the SVM classifier, which takes CNN features as input. Based on the predictions, Algorithm \ref{algorithm3} finally determines SLRB. For illustration convenience, the stride of the sliding window is enlarged to SCW.}
\label{pipeline_step1}
\end{figure}

Raw subtitle regions $RS$ bounded by the detected STBB and the left/right boundary of original frames are cropped from original frames. The size of $RS$ is $h_{s}\ \times \ W$, where $h_{s}$ represents subtitle height. Then, SLRB are detected in a sliding-window manner: a $h_{s}\ \times\  \left ( w-1 \right )$ window, a $h_{s}\ \times \ w$ window and a $h_{s}\ \times\ \left ( w+1 \right )$ window that respectively slide from left to right across $RS$ with stride 1 are adopted, where $w$ is the determined SCW. Then, every window region is classified as either text region or non-text region by the SVM classifier described in Section \ref{SVM}. Supposing that $a_{i}$ and $b_{i}$ respectively denote the left boundary position and the right boundary position of the $i$-th window region predicted as a text region, and there are $n$ window regions predicted as text regions. Algorithm \ref{algorithm3} is designed to merge overlapping window regions predicted as text regions together and subsequently determine the SLRB. According to the output $LeftBound$ and $RightBound$ of Algorithm \ref{algorithm3}, subtitle region $S$ is detected by further removing non-subtitle regions on two sides of $RS$. This process is illustrated in Fig.\ref{pipeline_step1}. The parameter $\beta$ of Algorithm \ref{algorithm3} is determined according to the resolution of videos. $\beta$ being too large would cause the real subtitle region to be easily connected with non-subtitle regions that are incorrectly predicted, while being too small, an integral sentence might be easily broken into pieces.

%
		%

\begin{algorithm}
\footnotesize
	\caption{SLRB determination}
	\label{algorithm3}
	\begin{algorithmic}[1]
	    \Require \begin{varwidth}[t]{\linewidth}
	        $n$ predicted text window regions $(a_{1},b_{1}),(a_{2},b_{2})\dots (a_{n},b_{n})$,\par
	        parameter $\beta$ controlling the maximum gap between two clauses separated by space,\par
	        the determined SCW $w$
	    \end{varwidth}
	    \Statex
	    \Ensure the left and the right boundarids of subtitle $\{LeftBound, RightBound\}$
        
        \Statex
        \State $i\gets 1,k\gets 1$
        \State $LeftCandidate\gets\varnothing,\ RightCandidate\gets\varnothing$
        \Statex
        \While {$i<n$} 
            \State $j\gets i+1$
            \State $right\gets b_{i}$
            \While {$j<=n$ \textbf{and} $a_{j}\leq right$}
                \State $right\gets {\rm max}(right,b_{j})$
                \State $j\gets j+1$
            \EndWhile
            \If {$j-i>3$}
                \If {$LeftCandidate=\varnothing$}
                    \State $RightCandidate[k]\gets right$
                    \State $LeftCandidate[k]\gets a_{i}$
                    \State $k\gets k+1$
                \Else
                    \If {$a_{i}\leq RightCandidate[k-1]+\beta \times w$ }
                        \State $RightCandidate[k-1]\gets right$
                    \Else
                        \State  $RightCandidate[k]\gets right$
                        \State $LeftCandidate[k]\gets a_{i}$
                        \State $k\gets k+1$
                    \EndIf
                \EndIf
                
            \EndIf
            \State $i\gets j$
       \EndWhile
       \Statex
       \State $Z\gets \mathop{\argmax}_{i}{(RightCandidate[i]-LeftCandidate[i])}$
       \State $LeftBound\gets LeftCandidate[Z]$
       \State $RightBound\gets RightCandidate[Z]$
	\end{algorithmic}
 \end{algorithm}


\subsection{Subtitle recognition}

Now that the subtitle region $S$ has been successfully detected, we will describe the proposed subtitle recognition scheme with three steps including sliding window based segmentation, window region recognition and dynamic programming determination.

\subsubsection{Sliding window based segmentation}

In order to recognize each single character in the subtitle, the subtitle region $S$ must be properly segmented (i.e. split the image text line into patches that each of which contains a single character). This step is challenging due to touching characters and the inherent structure of separation from the left and right sides of many East Asian characters. Unlike other methods where potential segmentation points must be determined precariously \cite{verma2003contour,Bai2014Chinese2,Bissacco2013PhotoOCR,Saidane2009The}, our method obviates this step since the SCW is  known, which is an inborn advantage of our system. Three sliding windows identical to those in the Section \ref{SLRB} are adopted again to slide from left to right across $S$ at stride one, and each window region is fed into the CNN ensemble for recognition.

\subsubsection{Window region recognition}

Given a window region ($a_{i}$, $b_{i}$), the softmax layer of each CNN model outputs the probability of each category, and categories whose probabilities are among the top 20 are reserved. Then, probabilities of these reserved categories are averaged across 10 CNN models. If the largest average probability is greater than a threshold (i.e. 0.2), candidate categories of ($a_{i}$, $b_{i}$) with the top 5 average probabilities will be recorded before moving to the next window position ($a_{i+1}$, $b_{i+1}$). Otherwise, the window region ($a_{i}$, $b_{i}$) would probably reside between two adjacent characters. In this case, it will be abandoned and the next window region ($a_{i+1}$, $b_{i+1}$) will be examined. Finally, those recorded 5 candidate categories whose probabilities are greater than 0.05 will be stored with their associated recognition probabilities $Rprob$ and the window position ($a_{i}$, $b_{i}$).


\subsubsection{Dynamic programming determination}
The final recognition results are determined by a dynamic programming algorithm. From the leftmost window ($a_{1}$, $b_{1}$) step by step all the way to the rightmost window ($a_{n}$, $b_{n}$), this algorithm builds the whole sentence by repeatedly appending the character in the next window position (i.e. $w-2$, $w-1$ or $w$ pixels rightward) to the previously recognized sentence. In each step from the window ($a_{i}$, $b_{i}$) to the next window ($a_{j}$, $b_{j}$), every previously recognized sentence that arrives to ($a_{j}$, $b_{j}$) is processed by a character based 3-gram language model. For every unique 3-gram word group consisting of the newly appended character and two former characters, a recognition probability $Rprob$ and a 3-gram language probability $Lscore$ are recorded, based on which the total score of the word group is calculated as:
\begin{equation}
\label{groupscore}
groupscore_{i,j}=\gamma\ \times \ \log(Lscore)+(1-\gamma)\ \times \ \log(Rprob),
\end{equation}
$\gamma$ is the proportion of the language score and the recognition score which is 0.3 in our experiment. Since the sliding window has three widths (i.e. $w-1$, $w$ and $w+1$), it is possible to obtain several identical word groups that arrive at $b_{j}$ but with different scores during the building process. Therefore, a pruning strategy that only reserves the word group with the highest score is applied to reduce the redundancy and improve the efficiency. The building process terminates when $b_{j}$ approaches the right boundary of the image, and the total score of the $k$-th possible sentence is:
\begin{equation}
\label{totalscore}
totalscore_{k}=\frac{\sum_{k}groupscore}{windows(k)},
\end{equation}
where $\sum_{k}groupscore$ represents the sum of all $groupscore$ in the $k$-th candidate sentence and $windows(k)$ represents the number of windows (i.e. characters) in the $k$-th candidate sentence. The sentence with the highest total score is selected as the final recognition result.

\section{Experiments}
\label{expr}

We conduct ample experiments to evaluate each component of the proposed system. The end-to-end performance of our system is also reported in this section.

\subsection{Dataset}
\label{human}
As listed in Table \ref{dataset}, an extensive dataset containing 1097 videos in Simplified Chinese, Traditional Chinese and Japanese is constructed. These videos exhibit a wide range of diversity in TV program genres, including talk shows, documentaries, news reports, etc.

STBBs of all videos and SLRBs of videos marked by $\dagger$ are annotated manually. As our recognition module is almost error-free, the recognition results of videos marked by $\dagger$ are annotated by a human annotator ``A'' on the basis of the outputs of the proposed system. The annotations obtained in this manner are regarded as ground truth. To test the quality of the ground truth annotations, we randomly select 400 frames containing 4494 characters from the already annotated frames and employ another two human annotators ``B'' and ``C'' to annotate these frames independently again. By comparing the annotations from ``B'' and ``C'', the final agreement on the result is reached, based on which the annotations from ``A'' are examined. The annotations from ``A'' achieve 99.8\% accuracy, indicating that the ground truth annotations are of high quality.

We also measure the human-level reading performance on these 400 frames. A human annotator ``D'' is employed to annotate these frames manually, and the annotations from ``D'' are examined based on the final agreement mentioned-above. The human-level reading performance is estimated by the performance of ``D'', of which the reading accuracy is 99.6\%.

\begin{table}[htb]
\centering
\footnotesize
\begin{tabular}{@{}lll@{}}
\toprule
Language            & \#Videos        & Resolution     \\ \midrule
Traditional Chinese & 1015 (40$^{\dagger}$) & 480$\times$320 \\
Traditional Chinese & 40              & 852$\times$480 \\
Simplified Chinese  & 40 (40$^{\dagger}$)   & 852$\times$480 \\
Japanese            & 2               & 480$\times$320 \\ \bottomrule
\end{tabular}
\caption{Our dataset configuration. All videos are utilized to evaluate the STBB detection module, while only videos marked by '$\dagger$' are randomly selected to evaluate the remaining modules and the end-to-end system.}
\label{dataset}
\end{table}

\subsection{Experiments on STBB and SCW detection}

In order to demonstrate the efficacy of our method, all videos in the dataset are selected for evaluation. In the experiment, the height of the vertical sliding window $h$ is optimized with regard to videos with 480 $\times$ 320 resolution and videos with 852 $\times$ 480 resolution respectively.

The CNN ensemble trained on synthetic data with random shift empowers our system with high robustness even if the STBB are not precisely detected. For this consideration, our evaluation method is defined as follows: the STBB of a video are detected correctly if
\begin{equation}
-3\leqslant T_{d}-T_{gt}\leqslant 2\ and\ -2\leqslant B_{d}-B_{gt}\leqslant 3,
\end{equation}
where $T_{d}$, $T_{gt}$, $B_{d}$ and $B_{gt}$ denote positions of detected top boundary, ground-truth top boundary, detected bottom boundary and ground-truth bottom boundary respectively.

We perform a series of tests to determine the optimal value of parameter $h$ (the height of the proposed vertical sliding window in Section \ref{CWT}). The input variables $w_{min}$, $w_{max}$ and $min\_height$ of Algorithm \ref{algorithm1} are set to 5, 40 and 12 respectively. Table \ref{parameter_h} shows the performance of our STBB detection module with regard to different $h$. The variable $h$ actually controls the trade-off between the STBB detection accuracy and the tolerability to noise. From our experiments, we observe that  when $h$ is too small, the histogram becomes more susceptible to background noise as well as strokes inside characters that do not reflect SCW. But $h$ being too large would compromise the STBB detection accuracy.

\begin{table}[htb]
\centering
\footnotesize
\begin{tabular}{@{}lllll@{}}
\toprule
\multicolumn{1}{c}{\begin{tabular}[c]{@{}c@{}}Video\\ resolution\end{tabular}} & \multicolumn{1}{c}{\begin{tabular}[c]{@{}c@{}}Number of\\ videos\end{tabular}} & $h$          & \multicolumn{1}{c}{\begin{tabular}[c]{@{}c@{}}Number of videos whose\\STBB are correctly detected\end{tabular}} & Recall            \\ \midrule
\multirow{4}{*}{480$\times$320}                                                & \multirow{4}{*}{1017}                                                          & 1            & 972                                                                                              & 95.6\%            \\
                                                                               &                                                                                & $\textbf{3}$ & $\textbf{980}$                                                                                   & $\textbf{96.4\%}$ \\
                                                                               &                                                                                & 5            & 951                                                                                              & 93.5\%            \\
                                                                               &                                                                                & 7            & 934                                                                                              & 91.8\%            \\ \midrule
\multirow{3}{*}{852$\times$480}                                                & \multirow{3}{*}{80}                                                            & 3            & 73                                                                                               & 91.3\%            \\
                                                                               &                                                                                & $\textbf{5}$ & $\textbf{75}$                                                                                    & $\textbf{93.8\%}$ \\
                                                                               &                                                                                & $\textbf{7}$ & $\textbf{75}$                                                                                    & $\textbf{93.8\%}$ \\ \bottomrule
\end{tabular}
\caption{Parameter $h$ optimization. STBB detection precision is not presented for the reason that false-positives are subsequently removed by the text/non-text classifier. Therefore, every video only has one final subtitle location. Note that the correctness of STBB determination always entail the correctness of SCW determination, hence only the former is reported.}
\label{parameter_h}
\end{table}


\subsection{Experiments on SLRB detection}

In this section, the performance of our SLRB detection module is evaluated against two baseline methods based on hand-engineered features: T-HOG \cite{Minetto2013T} and EOH-GSC \cite{wang2009new}. The input parameter $\beta$ of Algorithm \ref{algorithm3} is set to 0.7/2.5 for videos in 480 $\times$ 320/852 $\times$ 480 resolution respectively.

Our evaluation method is quite similar to the \emph{ICDAR'03 detection protocol} \cite{Lucas2003ICDAR}. Let $r$ denote the ground-truth SLRB, and $r'$ denote the corresponding detected SLRB. The average match $m_{ave}$ between all $r$ and $r'$ in a video is defined as twice the length of intersection divided by the sum of the lengths:
\begin{equation}
\label{ma}
m_{ave}\left (r,r'\right )=\frac{2\sum_{r\in E} L\left ( r\cap r' \right )}{\sum_{r\in E} \left ( L\left ( r \right )+L\left ( r' \right ) \right )},
\end{equation}
where $L(r)$ is the distance between a set of left and right boundaries and $E$ denotes all the ground-truth SLRBs in a video.

Table \ref{leftright} lists the statistics of $m_{ave}$ of 80 videos and shows the superiority of our CNN features over T-HOG \cite{Minetto2013T} and EOH-GSC \cite{wang2009new} features on the text/non-text classification task.


\begin{table}[]
\centering
\footnotesize
\begin{tabular}{@{}llll@{}}
\toprule
Language            & $\textbf{CNN features}$     & EOH-GSC \cite{wang2009new} & T-HOG \cite{Minetto2013T} \\ \midrule
Simplified Chinese  & $\textbf{99.4 $\pm$ 0.9\%}$ & 96.1 $\pm$ 2.5$\%$           & 91.7 $\pm$ 4.6$\%$          \\
Traditional Chinese & $\textbf{99.5 $\pm$ 0.4\%}$ & 96.8 $\pm$ 3.3$\%$           & 94.0 $\pm$ 5.1$\%$          \\ \bottomrule
\end{tabular}
\caption{The statistics of $m_{ave}$. We randomly select 80 videos (40 in Simplified Chinese and 40 in Traditional Chinese) whose STBBs are correctly determined for evaluation.}
\label{leftright}
\end{table}

\subsection{Experiments on subtitle recognition}

This section measures the performance of our character recognition module. For comparison, we test the same 80 videos in the previous section with Grayscale based Chinese Image Text Recognition (gCITR) \cite{Bai2014Chinese2} as well as another two commercial OCR software: ABBYY FineReader 12 \cite{ABBYY} and Microsoft OCR library \cite{msocr}. gCITR \cite{Bai2014Chinese2} is the previous state-of-the-art system for Simplified Chinese subtitle recognition, where 85.44\% word accuracy is achieved on another dataset. Besides, the performance of a single CNN is also reported in order to manifest the efficacy of the CNN ensemble.

The performance of our subtitle recognition module is evaluated by the word accuracy $W_{acc}$ that is defined as:
\begin{equation}
\label{wacc}
W_{acc}=\frac{N-E_{dis}}{N},
\end{equation}
here, $N$ is the number of ground-truth words and $E_{dis}$ represents \emph{Levenshtein edit distance} \cite{Levenshtein1965Binary} to change a recognized sentence into ground-truth.

\begin{table}[htb]
\centering
\tiny
\begin{tabular}{@{}llllllll@{}}
\toprule
TV programs & \#Videos & \#Words & ABBYY\cite{ABBYY} & gCITR\cite{Bai2014Chinese2} & MS OCR\cite{msocr} & Single CNN & \bf{CNN ensemble} \\ \midrule
HXLA        & 3        & 4630    & 52.4\%              & 78.5\%                        & 89.9\%               & 97.4\%     & \bf{99.7\%}       \\
CFZG        & 3        & 7711    & 78.7\%              & 91.8\%                        & 89.7\%               & 98.1\%     & \bf{99.7\%}       \\
ZGSY        & 3        & 8982    & 68.7\%              & 81.6\%                        & 85.8\%               & 98.5\%     & \bf{99.9\%}       \\
DA          & 2        & 3936    & 64.8\%              & 69.1\%                        & 89.0\%               & 97.7\%     & \bf{99.7\%}       \\
JXTZ        & 2        & 4682    & 66.8\%              & 70.3\%                        & 88.3\%               & 97.8\%     & \bf{99.6\%}       \\
FNMS        & 2        & 5681    & 68.3\%              & 87.7\%                        & 87.7\%               & 99.2\%     & \bf{99.8\%}       \\
JF          & 5        & 9299    & 54.3\%              & 75.8\%                        & 84.8\%               & 98.2\%     & \bf{99.3\%}       \\
KJL         & 2        & 3372    & 61.9\%              & 87.8\%                        & 61.3\%               & 98.0\%     & \bf{99.8\%}       \\
KXDG        & 1        & 2027    & 40.6\%              & 76.2\%                        & 56.3\%               & 97.5\%     & \bf{98.3\%}       \\
AQGY        & 2        & 4850    & 56.6\%              & 79.7\%                        & 56.9\%               & 94.3\%     & \bf{96.9\%}       \\
CCTVJS      & 2        & 3918    & 85.2\%              & 71.1\%                        & 82.6\%               & 96.2\%     & \bf{99.9\%}       \\
SDGJ        & 3        & 8700    & 67.0\%              & 83.2\%                        & 82.6\%               & 98.4\%     & \bf{99.9\%}       \\
DSGY        & 1        & 1872    & 68.9\%              & 31.4\%                        & 63.4\%               & 97.8\%     & \bf{99.0\%}       \\
JXX         & 1        & 3618    & 67.8\%              & 80.5\%                        & 71.7\%               & 97.7\%     & \bf{99.6\%}       \\
TTXS        & 1        & 2090    & 39.8\%              & 68.7\%                        & 86.3\%               & 96.7\%     & \bf{99.5\%}       \\
YSRS        & 3        & 8914    & 48.6\%              & 78.6\%                        & 80.8\%               & 98.1\%     & \bf{99.7\%}       \\
YST         & 2        & 4712    & 54.8\%              & 85.7\%                        & 85.9\%               & 97.1\%     & \bf{99.3\%}       \\
BBQN        & 1        & 2751    & 51.9\%              & 76.9\%                        & 76.8\%               & 96.1\%     & \bf{99.6\%}       \\
ZHDWM       & 1        & 1319    & 55.7\%              & 82.2\%                        & 52.4\%               & 95.9\%     & \bf{97.4\%}       \\ \midrule
Total       & 40       & 93064   &                     &                               &                      &            &                     \\
Average     &          &         & 62.0\%              & 79.4\%                        & 80.5\%               & 97.7\%     & \bf{99.4\%}       \\ \bottomrule
\end{tabular}
\caption{Word Accuracy of Simplified Chinese.}
\label{cs}
\end{table}

\begin{table}[htb]
\centering
\tiny
\begin{tabular}{@{}llllllll@{}}
\toprule
TV programs & \#Videos & \#Words & ABBYY\cite{ABBYY} & gCITR\cite{Bai2014Chinese2} & MS OCR\cite{msocr} & Single CNN & \bf{CNN ensemble} \\ \midrule
DXSLM       & 2        & 2024    & 62.8\%              & $-^{*}$                       & 86.8\%               & 98.2\%     & \bf{99.6\%}       \\
KXLL        & 10       & 11819   & 84.4\%              & $-^{*}$                       & 89.4\%               & 97.1\%     & \bf{99.5\%}       \\
NDXW        & 11       & 30683   & 38.3\%              & $-^{*}$                       & 47.9\%               & 96.7\%     & \bf{99.4\%}       \\
QJXTW       & 2        & 6245    & 34.4\%              & $-^{*}$                       & 61.9\%               & 97.9\%     & \bf{99.6\%}       \\
YXW         & 3        & 4361    & 54.0\%              & $-^{*}$                       & 63.4\%               & 97.5\%     & \bf{99.5\%}       \\
XWWW        & 4        & 10124   & 41.6\%              & $-^{*}$                       & 59.1\%               & 96.7\%     & \bf{99.5\%}       \\
XGD         & 2        & 5147    & 35.2\%              & $-^{*}$                       & 62.1\%               & 97.8\%     & \bf{99.4\%}       \\
XTWJY       & 2        & 4264    & 39.2\%              & $-^{*}$                       & 67.8\%               & 97.8\%     & \bf{99.6\%}       \\
XYZY        & 3        & 7603    & 93.2\%              & $-^{*}$                       & 85.4\%               & 97.3\%     & \bf{99.4\%}       \\
YHHS        & 1        & 2103    & 53.9\%              & $-^{*}$                       & 68.4\%               & 97.0\%     & \bf{99.6\%}       \\ \midrule
Total       & 40       & 84373   &                     &                               &                      &            &                     \\
Average     &          &         & 50.8\%              & $-^{*}$                       & 62.0\%               & 97.1\%     & \bf{99.4\%}       \\ \bottomrule
\end{tabular}
\caption{Word Accuracy of Traditional Chinese. $\textbf{*}$ gCITR \cite{Bai2014Chinese2} is not designed for Traditional Chinese.}
\label{trad}
\end{table}

Table \ref{cs} and Table \ref{trad} shows the performance of ABBYY \cite{ABBYY} , gCITR \cite{Bai2014Chinese2}, Microsoft OCR library \cite{msocr}, our single CNN and the CNN ensemble on the Simplified Chinese and Traditional Chinese text line recognition task. The performance of the proposed method exceeds other baselines by a large margin. In order to demonstrate the efficacy of our system on other languages, we also test it on two videos in Japanese, and an average 97.4\% $W_{acc}$ is achieved.

\subsection{End-to-end performance}
The same 80 videos in the previous section are selected for evaluating the end-to-end performance. Table \ref{end2end} compares the end-to-end performance of the proposed system with ABBYY \cite{ABBYY}, gCITR \cite{Bai2014Chinese2}, Microsoft OCR \cite{msocr}. 

\begin{table}[htb]
\centering
\footnotesize
\begin{tabular}{@{}lllll@{}}
\toprule
                    & ABBYY\cite{ABBYY} & gCITR\cite{Bai2014Chinese2} & MS OCR\cite{msocr} & $\textbf{Proposed}$  \\ \midrule
Simplified Chinese  & 60.7\%              & 78.1\%                        & 79.3\%               & $\textbf{98.2\%}$   \\
Traditional Chinese & 49.7\%              & -                             & 60.9\%               & $\textbf{98.3\%}$    \\ \bottomrule
\end{tabular}
\caption{End-to-end performance. Notice that three baselines take subtitle region detected by our system as input rather than raw video frames, as ABBYY \cite{ABBYY} and Microsoft OCR \cite{msocr} may generate many false detections on raw video frames and gCITR \cite{Bai2014Chinese2} can only perform text recognition.}
\label{end2end}
\end{table}

\section{Discussion}
\label{discuss}

\begin{figure}[htb]
\centering
\includegraphics[width=255pt]{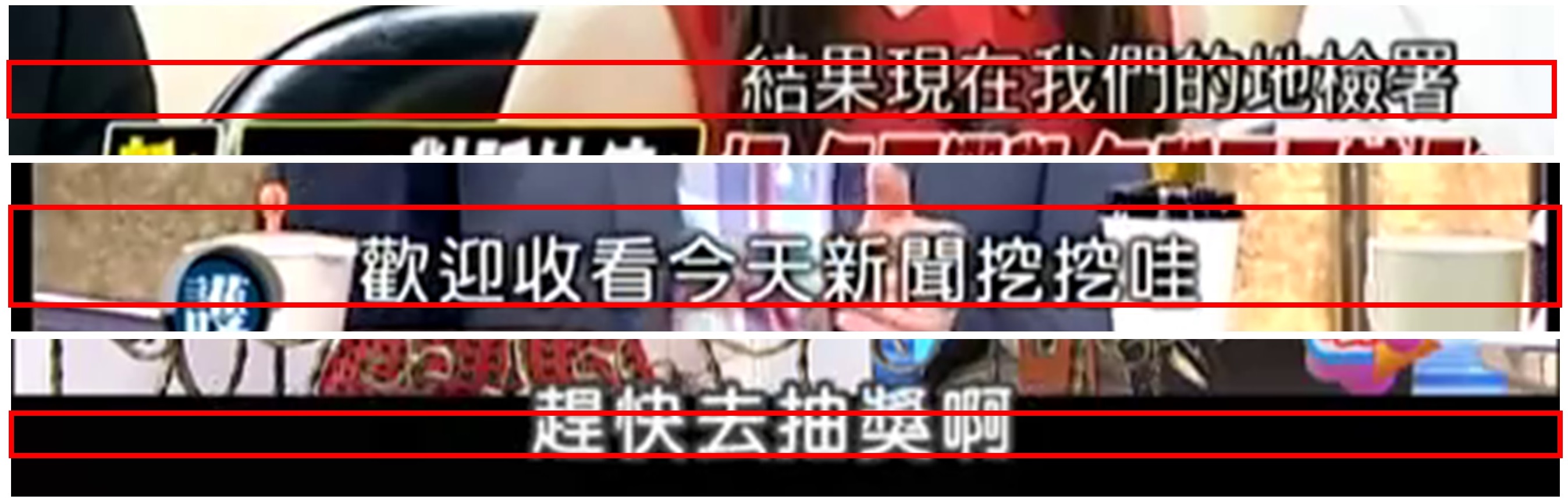}
\caption{Typical mistakes made by the STBB detection module. Red boxes denote the detected STBB.}
\label{tb_wrong}
\end{figure}

\begin{figure}[htb]
\centering
\includegraphics[width=255pt]{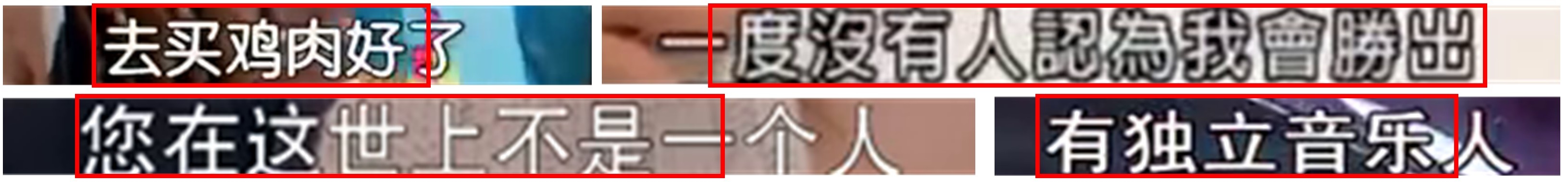}
\caption{Typical mistakes made by the SLRB detection module. Red boxes denote detected subtitle regions.}
\label{lefright_wrong}
\end{figure}

\begin{figure}[htb]
\centering
\includegraphics[width=255pt]{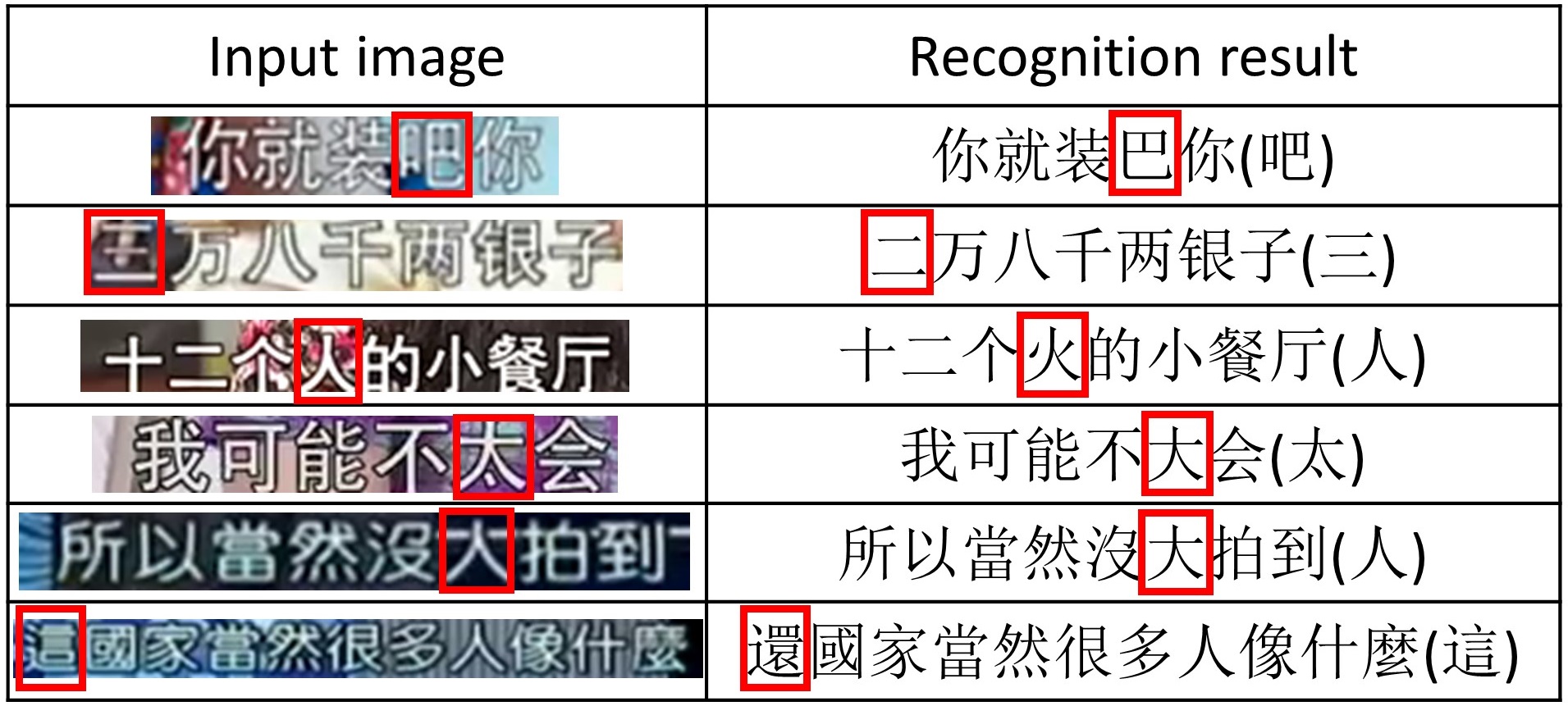}
\caption{Typical recognition mistakes made by the CNN ensemble. Red boxes mark the incorrectly recognized characters. The ground-truth characters are enclosed in parentheses.}
\label{recog_wrong}
\end{figure}

Although the STBB detection module has achieved competitive performance, there is still room for improvement. We observe that a majority of incorrectly detected STBBs locate near the ground-truth boundaries (Fig.\ref{tb_wrong}). Actually, more accurate boundary positions can be obtained if some regression methods like the one in \cite{jaderberg2016reading} are adopted. In the SLRB detection module, it is observed that specific characters are sporadically misclassified as non-texts. We find the strokes of these characters are all very sparse, which can be easily confused with edge or texture features at backgrounds (Fig.\ref{lefright_wrong}). Confusion and loss of radicals and strokes are two major mistakes made by the CNN character recognizer (Fig.\ref{recog_wrong}). Character categories that are misclassified more than three times are examined and the causes of the errors are scrutinized. We find that 45.5\% of the errors are caused by resemblances between two characters, 33.2\% are caused by cluttered backgrounds, 18.2\% are caused by the incorporation of the language model and 3.2\% are caused by large vertical shifts of characters.

\section{Conclusion}
\label{conc}
In this paper, we present an end-to-end subtitle text detection and recognition system specifically designed for videos with subtitles in East Asian languages. By applying CWT and integrating the sequence information throughout the video, we are able to detect STBB and SCW  simultaneously. This represents a departure from scene text detection problem where sophisticated methods are designed to detect texts in a single image. A CNN ensemble is leveraged to classify East Asian characters into thousands of categories. Our models are trained purely on synthetic data, which makes it possible for our system to be re-trained on other languages without requiring human labeling effort. Our system, as well as each module in it, compares favorably against existing methods on an extensive dataset. The near-human-level performance of our system qualifies it for practical application. For example, our system can provide accurate and reliable text labels for speech recognition researches, since video subtitles are synchronous with speech in videos.

In future work, this system will be tested on videos in Korean or other languages with consistent SCW. 

\section*{Acknowledgements}
This work is supported by Microsoft Research under the eHealth program, the Beijing Natural Science Foundation in China under Grant 4152033, the Technology and Innovation Commission of Shenzhen in China under Grant shengfagai2016-627, the Beijing Young Talent Project in China, the Fundamental Research Funds for the Central Universities of China under Grant SKLSDE-2015ZX-27 from the State Key Laboratory of Software Development Environment in Beihang University in China. We would like to thank Jinfeng Bai for conducting the gCITR baseline experiment.

\scriptsize
\begin{spacing}{1}
\bibliography{V1.bbl}
\bibliographystyle{elsarticle-num}
\end{spacing}
\end{document}